**BRAINCELL-AID: An Agentic AI Created Brain Cell Type Resource for Community Annotation**

Rongbin Li[1,*], Wenbo Chen[1,*], Zhao Li[1,*], Rodrigo Muñoz-Castañeda[2], Jinbo Li[1], Neha Maurya[1], Arnav Solanki[1], Huan He[3], Hanwen Xing[1], Meaghan Ramlakhan[1], Zachary Wise[1], Nelson Johansen[4], Zhuhao Wu[2], Hua Xu[3], Michael Hawrylycz[4], W. Jim Zheng[1,#]

1. McWilliams School of Biomedical Informatics, University of Texas Health Science Center at Houston, Houston, TX 77030, USA.
2. Appel Alzheimer's Disease Research Institute, Feil Family Brain and Mind Research Institute, Weill Cornell Medicine, New York, 10021, USA.
3. Department of Biomedical Informatics and Data Science, School of Medicine, Yale University, New Haven, CT, 06510, USA.
4. Allen Institute for Brain Science, Seattle, Washington, USA.

*Equal contribution
#Correspondence: wenjin.j.zheng@uth.tmc.edu

**SUMMARY**
Single-cell molecular transcriptomics and epigenomic data has transformed brain cell type identification, yet functionally annotating most putative types including novel or rare types remains challenging due to incomplete reference markers, and imperfect associations in the literature. While Large language models (LLMs) trained on biomedical literature show promise, their use is often limited by factual errors and imprecise biological reasoning.

We developed a novel multi-agent AI system—Brain Cell type Annotation and Integration using Distributed AI (BRAINCELL-AID: https://biodataai.uth.edu/BRAINCELL-AID)—for annotating brain cell types. BRAINCELL-AID includes a query agent to leverage fine-tuned LLMs trained on curated gene sets to learn principles of gene co-functionality. A literature agent and a retrieval-augmented generation (RAG) agent further enhance reliability by grounding annotations in peer-reviewed biomedical literature. Trained and evaluated on over 7,000 gene sets from MSigDB, BRAINCELL-AID achieved a record high concordance with biological ground truth: 77% of mouse and 74% of human gene sets contained annotations highly relevant to the known biology, when assisted by the RAG component. BRAINCELL-AID currently provides annotations for over 20,000 brain cell type–specific marker gene sets derived from single-cell RNA-seq data across 5,300+ brain cell type clusters spanning the entire mouse brain, integrating multiple biological signatures and contextual information. BRAINCELL-AID's annotation search enables identification of Basal Ganglia-related cell types with neurologically meaningful descriptions. In contrast, traditional gene set enrichment analysis lacks identifiable cell type labels and only yields generic functional terms for the Basal Ganglia cell types identified by BRAINCELL-AID.

BRAINCELL-AID enables novel insights into brain cell function by identifying region-specific gene co-expression patterns and inferring functional roles of gene ensembles, associating gene sets with novel regulatory factors. BRAINCELL-AID predicts new regulatory function, such as dual-transmitter signaling through the spatially restricted co-expression of Slc6a3, Gtf2a1l, and Aldh1a1 across Substantia Nigra (SN), Ventral Tegmental Area (VTA), and Midbrain Raphe Nuclei (RAmb), modulating dopaminergic signaling with potential GABAergic co-transmission and identifies similar molecular signatures in the transcriptional regulartor RAmb. This highlights BRAINCELL-AID's ability to generate testable hypotheses about gene function and circuit mechanisms in understudied regions, supporting scalable, interpretable discovery in neuroscience.

To ensure transparency and continuing expert validation, the accurate, literature-backed BRAINCELL-AID annotations are provided as a shared resource for neuroscience community to collaboratively evaluate, refine, analyze and annotate brain cell types (https://biodataai.uth.edu/BRAINCELL-AID), enabling human-AI collaboration. This hybrid framework lays the groundwork in application to other neural systems for high-quality annotation of comprehensive, expert-informed cell atlases across species.

## INTRODUCTION

The human brain, composed of over 85 billion neurons, a comparable number of non-neuronal cells, and interconnected by more than a trillion synapses, is the most complex known biological systems. Understanding the identity and function of these cells is a critical step toward deciphering brain function. To address this complexity, the NIH BRAIN Initiative Cell Atlas Network (BICAN) (RRID:SCR_022794) has fostered a collaborative research community focused on mapping brain cell types, connections, and functions across time and space in both human and non-human primates. Leveraging advances in single-cell RNA sequencing (scRNA-seq), epigenomic and multiomic techniques, BICAN profiles brain cell diversity at unprecedented resolution, revealing transcriptomic signatures that reflect underlying biological processes.[1-4] Through these efforts, BICAN has identified a broad array of brain cell types based on transcriptomic profiles. For example, Yao et al.[3] reported 5,322 distinct cell clusters in the mouse brain, identified through transcriptome data, gene regulation patterns, and anatomical localization, culminating in a high-resolution mouse brain cell atlas. Similarly, 3,313 brain cell clusters have been characterized in the human brain.[5]

These cell clusters are distinguished by unique marker gene sets, which offer the potential to identify insights into their identity and potential functions. Yao et al. categorized marker genes into three types: "Cluster Combo" genes from scRNA-seq data, "MERFISH" genes from spatial transcriptomics, and "TF" genes derived from transcription factor analysis. In addition, the top differentially expressed genes—treated as additional gene sets—can provide novel insights into brain cell function. Collectively, these gene sets define the molecular signatures of brain cell clusters and serve as a foundation for cell type classification.

Although these gene sets are ubiquitous in molecular neuroscience, most marker gene sets remain poorly annotated. Traditional gene set annotation tools, such as Gene Set Enrichment Analysis (GSEA) (RRID:SCR_003199),[6] rely on predefined annotation databases, which are often incomplete, generic, lack of detail, or referencing outdated literature. These tools struggle to capture the diversity and novelty of brain cell populations, limiting their utility in discovering new gene functions, interactions, and pathways.[7-11]To date, nearly 90% of mouse brain cell clusters lack clear and relevant biological interpretation, primarily due to the complexity of the brain and limited functional characterization of brain-expressed genes which presents a major barrier to defining cell types and understanding their roles.

Poorly annotated brain marker gene sets and cell types also pose a significant barrier to achieving the FAIR data principles (Findability, Accessibility, Interoperability, and Reusability). Without comprehensive and structured annotation, datasets become difficult to locate using keyword searches or other biologically meaningful queries, thereby undermining findability and accessibility. Furthermore, the absence of controlled vocabularies and standardized terminologies, such as the Gene Ontology (GO) (RRID:SCR_002811)[12-15] for gene sets, renders data non-interoperable across systems, studies and species. Finally, inadequate or superficial annotation fails to convey meaningful biological context, severely limiting the reusability of the data. Collectively, these deficits hinder the integration, comparison, and application of brain cell data across the research community, reducing its scientific value and impact.

LLMs offer a promising alternative to annotate brain cell marker gene sets.[16] Models like Llama 4,[17] trained on over 22 trillion tokens, can annotate gene sets and cell types without depending solely on curated databases.[18-23] However, current LLM-based approaches face key limitations: they often struggle to accurately contextualize structured biological terms (e.g., Gene Ontology concepts) and are susceptible to hallucinations and unverifiable outputs.[24-26]

To address these challenges, we developed BRAINCELL-AID (Brain Cell Type Annotation and Integration using Distributed AI) (RRID:SCR_027398), an AI agent-based system for annotating brain cell types. The process begins with marker gene set annotation by fine-tuned LLMs and integrates additional contextual information—such as anatomical region, neurotransmitter type, and neurofunctional properties—to generate a comprehensive cell type description. To enhance accuracy and minimize hallucinations, BRAINCELL-AID incorporates retrieval-augmented generation (RAG),[27] which grounds annotations in current relevant PubMed literature.

BRAINCELL-AID demonstrated strong performance, with 77% of mouse gene sets and 74% of human gene sets containing at least one annotation that was biologically relevant to the ground truth in its top 5 predictions during testing. These results indicate that BRAINCELL-AID can provide high-quality initial annotations of brain cell types, serving as a valuable community resource for efficient collaborative refinement and validation by human experts, ultimately leading to highly accurate brain cell type annotations. Using this new framework, we systematically annotated marker gene sets and cell types across the entire adult mouse brain as characterized in the Allen Brain Cell Atlas (ABC Atlas) (RRID:SCR_024440), to provide a scalable and reliable resource for the neuroscience community. To support community-driven annotation, BRAINCELL-AID includes interactive features that enable users to evaluate, edit, and contribute annotations based on domain expertise. Through tools for edition, validation, and submission, the platform fosters collaboration and serves as a foundational resource for the research community to advance brain cell type research.

## RESULTS

### Data

As a collaborative effort among neuroscientists, computational biologists, and software engineers, BICAN is producing a comprehensive brain atlas for humans and 12 other species, using six investigative modalities and ten experimental techniques (BICAN Website: https://www.portal.brain-bican.org/). This large-scale initiative has generated extensive raw data, freely shared with the research community through repositories such as NeMOarchive (RRID:SCR_016152),[28] Brain Image Library (BIL) (RRID:SCR_017272),[29] and Distributed Archives for Neurophysiology Data Integration (DANDI) (RRID:SCR_017571).[30] Metadata, processed data, protocols, and tools are openly accessible via platforms like GitHub (RRID:SCR_002630).

A central resource emerging from this work is the Allen Brain Cell (ABC) Atlas, which enables in-depth exploration of brain cell diversity and function across species and modalities. In this study, we focus on the Mouse Whole-Brain Cell Type Atlas spanning the entire brain (**Table 1**).[3] This atlas integrates single-cell transcriptomic and spatial data from over 4 million cells, resulting in a high-resolution taxonomy of 5,322 distinct brain cell clusters. These clusters are hierarchically organized into 34 classes, 338 subclasses, and 1,201 supertypes, providing a detailed map of brain cell diversity.[3] Each cluster is annotated by anatomical location, neurotransmitter identity, and marker gene sets derived from three distinct modalities: 1) Cluster Combo genes (from scRNA-seq); 2) MERFISH genes (from spatial transcriptomics); and 3) TF genes (from transcription factor analysis). While cell types and functions are relatively well-characterized at the supercluster level, deeper levels present increasing complexity, with many subclusters lacking functional annotation. Moreover, the predefined marker gene sets are modality-specific and often lack comprehensive functional interpretation, as discussed in the Introduction.

**Table 1. Mouse brain cell atlas data used in this study.**

| | |
|---|---|
| **Source** | Yao *et al.* |
| **Species** | Mouse |
| **Brain Region** | Whole adult mouse brain |
| **Number of cells** | 4 million cells |
| **Sequencing Technologies** | Single-cell RNA sequencing<br>• 10x Genomics Chromium v2<br>• 10x Genomics Chromium v3<br>• 10x Multiome snRNA-seq<br>MERFISH |
| **Structure of Taxonomy** | 4 levels of classification |
| **Cell clusters at each level** | -34 classes<br>-338 subclasses<br>-1,201 supertypes<br>-5,322 clusters |
| **Average number of cells/cluster** | 760 |
| **Number of marker gene Sets** | 5322 Cluster_Combo<br>5322 TF<br>5322 MERFISH |
| **Additional marker gene sets** | 5309 Top 20 DEG (This study) |

To address these limitations, we generated an additional marker gene set based on the top 20 differentially expressed genes (DEGs) for the 5309 out of 5322 clusters. While these DEGs will intersect with genes used in determining the original clusters, they complement the existing marker sets and strengthens cluster characterization by capturing additional functional signals (**Methods**). These DEGs provide a foundation for annotating poorly characterized clusters. For clarity, all the analysis in this study will be conducted at the cluster level, unless specified otherwise.

### Agentic workflow and goal overview

An AI agent is an intelligent entity capable of achieving specific goals using LLMs and other AI methods. Agents operate collaboratively within a system, enabling modular, scalable solutions that handle complex tasks through decomposition, planning, and interaction.[31,32]

BRAINCELL-AID is composed of three coordinated agents, forming an agentic workflow for gene set and brain cell annotation (**Figure 1**):

- The Query Agent takes a gene set as input and uses a fine-tuned LLM to generate an initial description of its collective biological function.
- The Literature Agent retrieves relevant publications based on the gene set and preliminary annotation.
- The RAG (Retrieval-Augmented Generation) Agent uses the retrieved literature to refine the initial annotation, grounding it in biomedical evidence.

This multi-agent system enables end-to-end annotation through prediction, literature retrieval, and refinement, automating the annotation process to support future community curation.

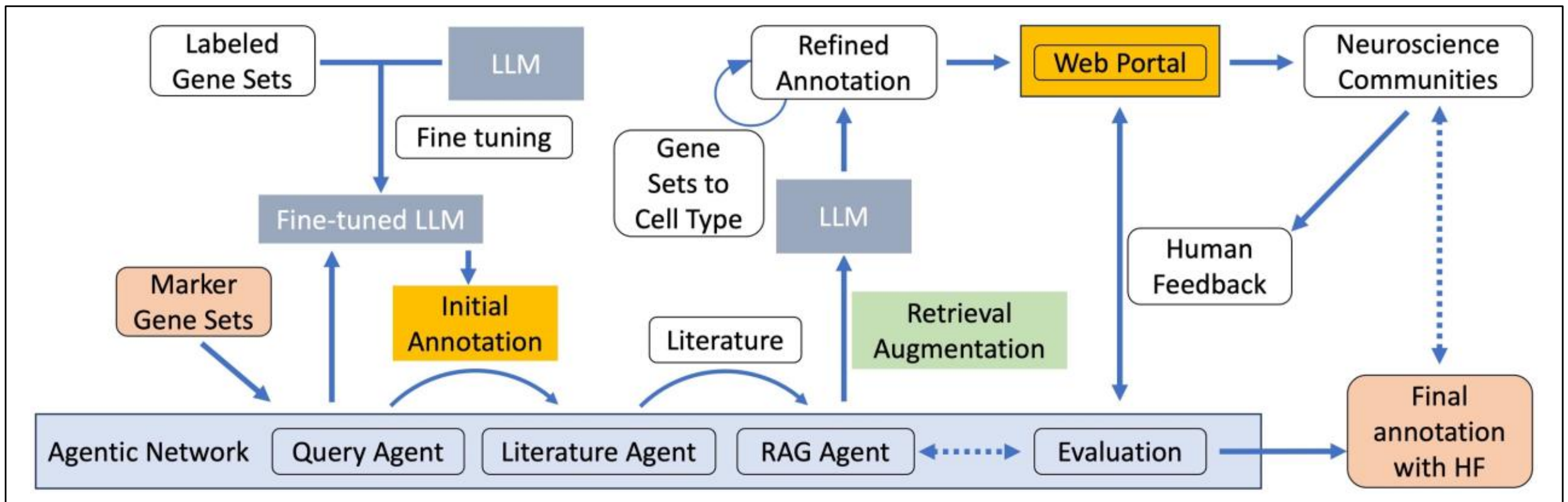

**Figure 1**. **Agentic network workflow and goal overview for comprehensive brain cell type annotation**. The agentic workflow of BRAINCELL-AID, detailing the components and workflow that an input marker gene set will undergo to obtain the final annotation.

## Gene set annotation – the Query Agent with fine-tuned LLMs

Annotating brain cell gene sets is the first step to understanding brain cell types and their functions, but current tools are limited. LLMs offer a transformative opportunity to overcome this challenge. Trained on the full corpus of biomedical literature, LLMs have the potential to interpret complex biological concepts, annotate datasets, explain experimental findings, and even propose hypotheses or therapeutic insights. However, most general-purpose LLMs lack a deep understanding of gene co-functionality, often producing imprecise or hallucinated annotations when applied to biological data.[33,34] This limitation underscores the need for more targeted training to enhance biological relevance and reliability.

To address this, we fine-tuned LLMs using over 7,000 well-annotated gene sets from the Molecular Signatures Database (MSigDB) (RRID:SCR_016863),[35] enabling the models to better capture gene co-functionality (**Figure 2A; Methods**). Leveraging ontology narration, or ontology verbalization, via GPT-4,[36] we transform structured ontology terms into natural language descriptions while preserving their meaning (**Methods**).[37] This aligns better with LLMs' language-based pretraining, enabling them to more effectively learn biological concepts. This strategy, termed GPTON (Generative Pretrained Transformer enhanced with Ontology Narration),[38] significantly enhances the model's ability to generate accurate and coherent gene set annotations, highlighting the effectiveness of fine-tuning with high-quality biological data.

Our fine-tuned models achieved substantial performance gains: for human gene sets, ROUGE-1[39] scores improved by 14% over GeneAgent[33] and 20% over GPT-4[34] (**Figure 2B**). The Llama human and mouse gene sets (**Figures 2B, 2C, and S1; Tables S1–S4**). Notably, models fine-tuned using verbalized ontology terms outperformed those trained directly on GO labels, highlighting the benefit of natural language alignment. We then selected the Llama 3 8B model for further evaluation due to its strong performance and compact size. The model showed stable performance across five-fold cross-validation, with consistent scores across all evaluation metrics (**Figures 2D and S2; Table S5**).

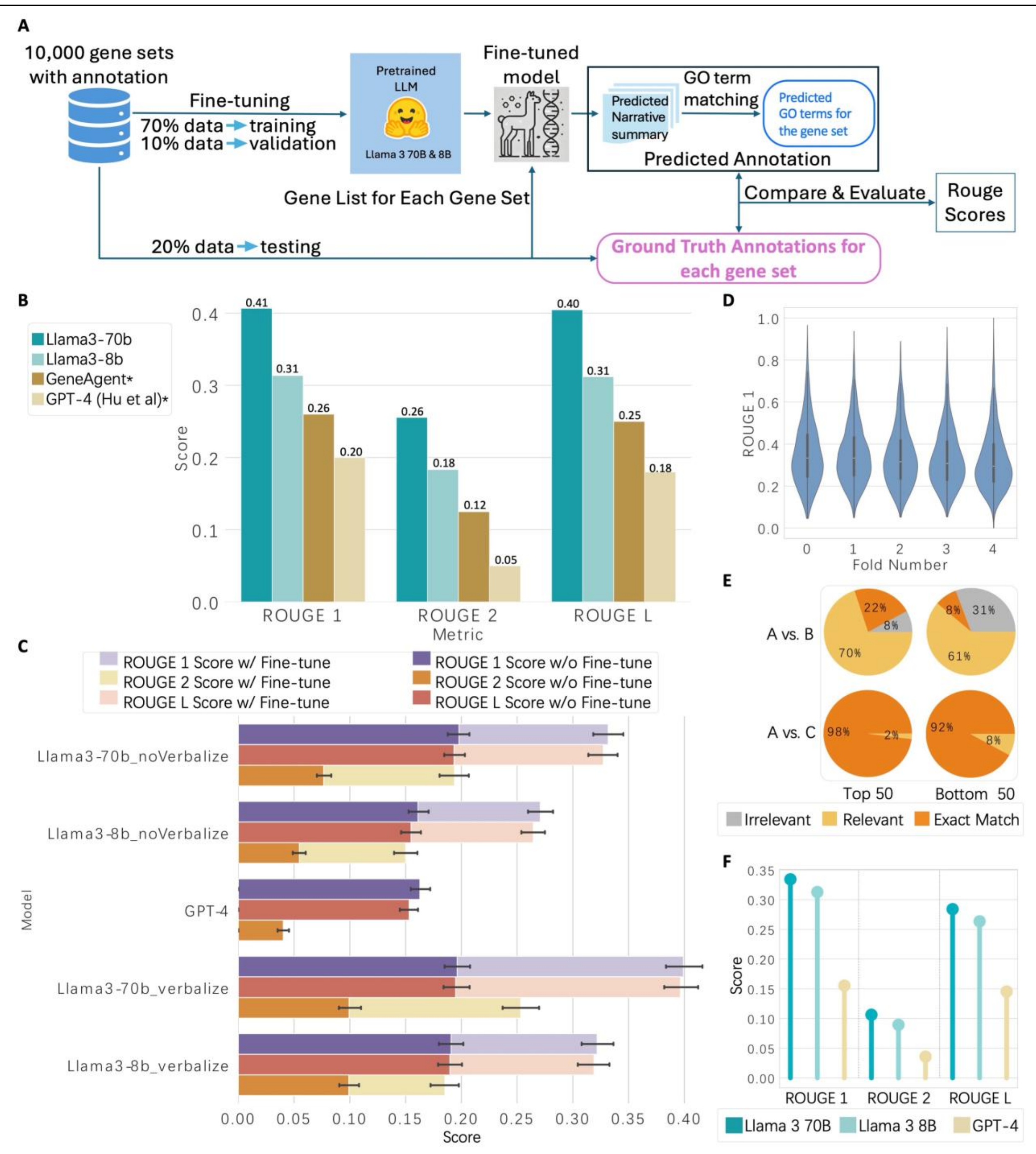

**Figure 2**. **Fine-tuning LLMs significantly improves gene set annotation.** (A) Workflow for fine-tuning LLMs using labeled gene sets. (B) ROUGE scores of the fine-tuned LLMs, including Llama 3 70B and 8B fine-tuned with verbalized GO terms, compared to GeneAgent and GPT-4 for human gene sets annotation. (C) ROUGE scores of Llama 3 models fine-tuned with verbalized ontology terms (verbalize) versus original ontology terms (noVerbalize), along with GPT-4 model for direct GO term generation for mouse gene sets annotation. The error bars represent 95% confidence intervals. (D) Consistent performance of the fine-tuned Llama 3 8B model evaluated through 5-fold cross-validation on mouse gene sets. The thick lines represent interquartile ranges, and thin lines are 1.5x interquartile ranges. (E) Human evaluation of the correct and relevant GO term predictions (A vs. B row) and consistent GO term verbalization (A vs. C row) in 100 selected gene sets. (F) Performance of fine-tuned Llama 3 70B and 8B models on the mouse Reactome dataset.

Two bioinformaticians manually evaluated the alignment between predicted and original GO terms (**Methods**) to assess annotation accuracy. 100 gene sets were sampled and ranked by the ROUGE-1 scores achieved by their corresponding GPTON predictions. For the top 50 gene sets, 92% of predicted terms exactly matched or were highly relevant to the originals; for the bottom 50, the rate was 69% (**Figure 2E**). We also assessed GPT-4 verbalizations of GO terms, which aligned meaningfully with the original terms in 98% (top 50) and 92% (bottom 50) of cases. These evaluations confirm the model's reliability in reproducing accurate biological meanings.

To test generalizability and robustness, we evaluated the models' performance on the Reactome datasets that are not used for the model training. The Reactome data has 1,670 human and 1,247 mouse gene sets respectively (**Methods**). The fine-tuned models achieved ROUGE-1[39] scores of 0.32 (human) and 0.33 (mouse), significantly outperforming GPT-4 (0.16 for human and 0.13 for mouse) (**Figure 2F; Table S6**). This improvement (**Figure S3; Table S6**) demonstrates strong generalization, even to datasets unseen during training. Performance also scaled positively with training data size, suggesting further gains are possible with expanded training sets (**Figure S4; Table S7**). These results showcase the reliability and robustness of GPTON's approach.

In BRAINCELL-AID, we implemented our AI Query Agent to access GPTON based on Llama 3.0 70B model—the largest model our computing system can support—which was fine-tuned exclusively on mouse gene sets. To enhance its relevance for neuroscience applications, we designed prompts to guide the model toward neurological functions (**Methods**). This tailored approach enabled accurate and context-specific initial annotations for cell clusters in the ABC Atlas, contributing to the construction of a high-resolution, biologically grounded brain cell taxonomy.

**Refine Annotation via Retrieval Augmented Generation (RAG) by Literature Agent and RAG Agent**

We developed fine-tuned LLMs that generate coherent and biologically meaningful annotations for gene sets (**Figure 2**). However, like all LLMs, it is prone to hallucinations. To address this, we implemented an 11-step RAG strategy that grounds annotations in published literature (**Figure 3A; Methods**). A central feature is the Literature Agent, which retrieves and ranks relevant PubMed abstracts to support both genes and inferred functions. By combining two complementary sets—TopPM (most relevant abstracts overall) and TopGene (one top abstract per gene)—our approach ensures both depth and breadth, enhancing accuracy, completeness, and reliability of gene set annotations (**Figure 3A, Step 1-3; Methods**).

PubMed abstracts curated by the Literature Agent are used by a comprehensive RAG agent to improve the accuracy and reliability of gene set annotations (**Figure 3A, Step 4-7; Methods**). Using the Llama 4 405B model, the agent integrates gene lists, initial LLM summaries, literature evidence, and cell context to refine annotations. It then produces concise two-sentence summaries grounded in verified literature, separately for TopPM and TopGene sets. Finally, each summary is linked to relevant Gene Ontology terms, ensuring structured, biologically meaningful interpretation and supporting more accurate, evidence-based cell type classification.

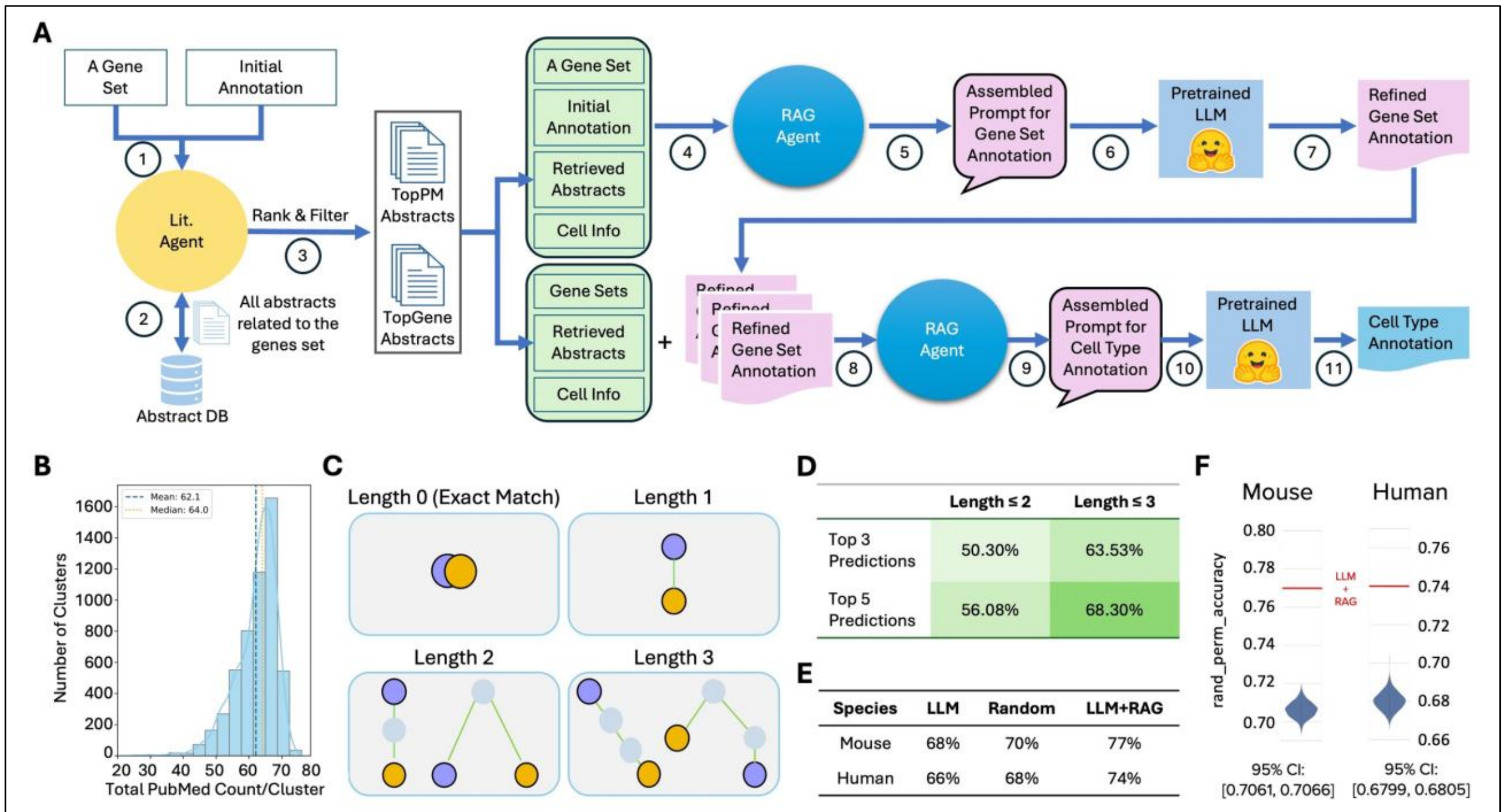

**Figure 3. BRAINCELL-AID RAG workflow for comprehensive marker gene set and brain cell type annotation.** (A) An 11-step Retrieval-Augmented Generation process retrieves relevant literature (Step 1-3) and integrates transcriptomic signatures, literature, cell-specific information (e.g., anatomical location, neurotransmitters) to annotate brain marker gene sets (Step 4-7) and cell types (Step 8-11). (B) A histogram shows that on average, each cell type has more than 60 relevant PubMed abstracts available for meaningful RAG. (C) GO graphs are used to evaluate annotation accuracy. (D) A fine-tuned large language model alone achieves 68% annotation accuracy on gene sets as evaluated by MSigDB data sets. (E) Incorporating RAG improves annotation accuracy to 77% for mouse and 74% for human gene sets. (F) Bar plots compare annotation accuracy of LLM-enhanced annotation with and without RAG, and against ten randomly selected GO terms. A violin plot displays the distribution of annotation accuracies across 1,000 random GO term samples for both human and mouse gene sets—showing significantly lower accuracy than RAG predictions.

We evaluate BRAINCELL-AID's performance using the structure of the GO graph. For gene set annotation, the query agent with fine-tuned LLM identified exact or highly relevant terms—defined as within three edges of the ground truth (**Figure 3C**)—in over 68% of mouse gene sets within the top five predictions (**Figure 3D**). This objective, topology-based evaluation reinforces the model's utility for real-world gene set annotation.

To evaluate the effectiveness of the RAG agent, we applied the same assessment framework. For each gene set in the human and mouse datasets, we identified the top 5 GO term predictions from three annotation sources: the original GPTON output, the RAG-refined TopPM, and TopGene prompts—resulting in 15 GO term predictions per gene set.

We then used the GO graph topology to assess annotation accuracy, measuring the proportion of gene sets where at least one of the 15 predicted GO terms was highly relevant to the ground truth. The RAG-enhanced approach led to a notable improvement: accuracy increased from 68% to 77% for mouse, and from 66% to 74% for human gene sets (**Figure 3E**).

To ensure these improvements weren't simply due to having more GO term predictions, we conducted a control experiment. We replaced the 10 RAG-predicted GO terms with randomly sampled terms from the Biological Process branch and repeated the process 1,000 times. The average accuracy from these random sets reached only 70% for mouse and 68% for human datasets (**Figure 3E**).

A one-sample t-test confirmed that the accuracy gains from RAG-generated GO terms were statistically significant compared to the random controls (**Figure 3F**), validating that the improvement was not due to chance.

These results demonstrate that the RAG methodology meaningfully enhances GO term prediction accuracy and supports more precise biological interpretation of gene sets. This highlights its value as a robust tool for advancing functional genomics research.

**Annotate marker gene sets for Brain Cell Atlas**

BRAINCELL-AID was applied to annotate comprehensive brain cell atlas datasets for mouse[3] (**Table 2; Methods**). This atlas offers fine-grained taxonomies—organized into classes, subclasses, supertypes, and clusters—that map the cellular composition of the brain[3]. However, despite their structural details in taxonomy, they lack consistent and comprehensive functional annotations. The atlas also provided three marker gene sets: Combo Markers, Transcription Factor Markers, and MERFISH Markers. While these gene sets offer insights into potential cell functions, they are often limited in scope and do not fully capture the functional diversity of brain cell types. To address this, we identified the top 20 differentially expressed genes for each cell cluster to generate richer, functionally informative gene sets (**Methods**).

**Table 2. Cell type cluster counts of the gene sets and cell types annotated by BRAINCELL-AID.** Total cell type cluster: 5322. BG: gene sets or cell type clusters with the keyword 'Basal Ganglia' in their annotation. Numbers in the parentheses include BG appeared in the Cell Cluster Annotation Summary.

| | | BRAINCELL-AID Annotation | | | | GSEA Annotation | | | |
|---|---|---|---|---|---|---|---|---|---|
| | | Contributing Genes | | | BG | Contributing Genes | | | BG |
| | | >0 | >1 | >2 | | >0 | >1 | >2 | |
| Gene Set Summary | Cluster_Combo | 5041 | 5012 | 4450 | 23(48) | 5245 | 2209 | 159 | 0 |
| | MERFISH | 5300 | 5299 | 5239 | 39(64) | 5244 | 3753 | 668 | 0 |
| | TF | 5316 | 5314 | 5229 | 168(187) | 5299 | 4972 | 3062 | 0 |
| | Top20 | 5308 | 5308 | 5308 | 56(62) | 4686 | 4686 | 4539 | 0 |
| | Total | 20965 | 20933 | 20226 | 246 | 20474 | 15620 | 8428 | 0 |
| Cell Type Summary | | 5322 | 5322 | 5315 | 33 | 5132 | 5131 | 5099 | 0 |

Using BRAINCELL-AID, we analyzed these four types of marker gene sets for each mouse cell cluster. BRAINCELL-AID, powered by generative language models and supported by literature retrieval, successfully generated biologically grounded, literature-backed narrative annotations for these brain cell clusters (**Table 2; Methods**). In total, BRAINCELL-AID annotated all 21,275 gene sets across 5,322 mouse brain cell clusters, with 20,965 of these gene sets supported by gene-relevant literature evidence (**Table 2**).

For comparison, we applied GSEA to these gene sets, which produced annotations for 20,474 clusters (**Table 2; Methods**). However, 15,620 of these annotations were based on only two genes, and 8,428 relied on just three genes. In contrast, BRAINCELL-AID generated annotations for 20,226 gene sets using three or more genes. Since effective cell type annotation should reflect key cellular functions derived from co-expressed genes, the sparse annotations produced by GSEA are insufficient to capture the functional identity of the cells (See analysis and comparison below).

Manual comparisons between annotations generated by gene set alone and those refined by RAG show that RAG consistently improves both specificity and biological relevance. For example, for the cluster_combo gene set *Slc6a3, Satb2, Bmp3, Sln* in Cluster 1571, annotation based on this gene set alone initially produced a generic annotation:

> *"Mechanisms that decrease or inhibit the movement of single-atom anions across cell membranes..."*.

After applying RAG with the TopPM prompt, the annotation was refined to:

> *"The Slc6a3, Satb2, Bmp3, Sln gene module is associated with a dopaminergic neuron subtype in the Arcuate hypothalamic nucleus-Periventricular hypothalamic nucleus, intermediate part, potentially influencing dopamine neurotransmission and regulation."*

This shift from a vague physicochemical description to a precise, functionally grounded narrative highlights the impact of relevant literature retrieved by the Literature Search Agent, enabling a deeper interpretation of gene function.

These annotated gene sets form a robust foundation for understanding cell-type-specific functions and building a detailed, functionally annotated brain cell atlas across species.

**Comprehensive brain cell type annotation**

We generated comprehensive brain cell type annotations by integrating transcriptomic, regulatory, and spatial signatures with anatomical, neurochemical, and metadata features, then grounding them in supporting literature (**Figure 3A, Step 8-11; Methods**). Using the Llama 4 405B model, we synthesized these inputs into coherent, biologically meaningful descriptions for each cell cluster (**Table 2; Methods**). This approach moves beyond transcriptomic profiling alone, providing rich and interpretable annotations that connect molecular, functional, and spatial context—offering a more complete view of brain cell identity and function.

For comparison, we also combined all marker genes for each cell cluster and performed collective GSEA to provide a brain cell type summary. Given the larger gene sets—and the presence of many well-studied genes in TF and MERFISH data—it is not surprising that most clusters received annotations through this approach (**Table 2; Methods**). However, these annotations tend to be vague, superficial, and generic, lacking relevance to the neurological functions of brain cells (see section below).

**BRAINCELL-AID brings enhanced scientific annotations and findings**
BRAINCELL-AID offers detailed, literature-based annotations with comprehensive coverage of biological functions for each brain cell type. Its powerful search capabilities allow users to quickly identify brain cells associated with specific biological or physiological functions, providing a broad and integrated view of these cells across the entire brain. This level of insight is unmatched by current approaches like GSEA.

We queried the BRAINCELL-AID database for cell types associated with the Basal Ganglia by searching annotation containing the keyword "Basal Ganglia". This search returned 21 classes, 103 subclasses, 175 supertypes, and 246 brain cell clusters (**Table 2**). **Figure 4A** shows the hierarchical taxonomy of these Basal Ganglia–related clusters, while **Figure 4B** maps them to the broader BICAN Brain Cell Atlas Taxonomy at the class, subclass, and supertype levels.

We also searched the GSEA-derived annotations for the keyword "Basal Ganglia" but found no matching clusters—highlighting a fundamental limitation of GSEA's reliance on predefined terms. In contrast, BRAINCELL-AID's annotations are enriched with relevant neurological context, including supporting literature and gene-level evidence, offering users deeper insight and interpretability.

Beyond limited keyword matching, GSEA annotations are often overly generic and uninformative. To compare annotation quality, we performed word cloud analyses (**Methods**) on Cell Cluster Annotation Summary using three sources: (1) BRAINCELL-AID annotations for all mouse brain cell clusters; (2) BRAINCELL-AID annotations for clusters containing "Basal Ganglia"; and (3) GSEA annotations for the same set of clusters identified by BRAINCELL-AID (since GSEA itself failed to detect Basal Ganglia–specific clusters).

As shown in **Figure 4C**, BRAINCELL-AID annotations capture rich neuronal characteristics—such as neurotransmission, glutamatergic, GABAergic, synaptic, neural, etc. For the Basal Ganglia clusters specifically (**Figure 4D**), BRAINCELL-AID annotations include highly relevant terms like dopamine and motor, distinguishing them from general brain cell features. In contrast, **Figure 4E** demonstrates that GSEA annotations are vague, lacking brain-specific or Basal Ganglia–relevant terms, even though all analyzed cell clusters are derived from brain tissue. Further details can be found in **Methods: Word Cloud Analysis, Table S8** and **Figure S5**.

In summary, BRAINCELL-AID provides deeper, more targeted insights into brain cell function, offering a significant advantage over traditional enrichment-based methods like GSEA.

BRAINCELL-AID also demonstrates its ability to accurately capture the known biological roles and activation states of non-neuronal cells. We analyzed Clusters 5312, 5313, and 5314 to validate their glial cell annotations on microglia and border-associated macrophages (BAMs).

Cluster 5312 was identified as a microglial population based on canonical markers such as *Tmem119*, *Sall1*, and *Cx3cr1*.[40] BRAINCELL-AID's annotation summary correctly highlighted key microglial functions, including immune surveillance, synaptic pruning, and homeostatic maintenance. Interestingly, expression of *Slc1a3*, atypical for microglia, was

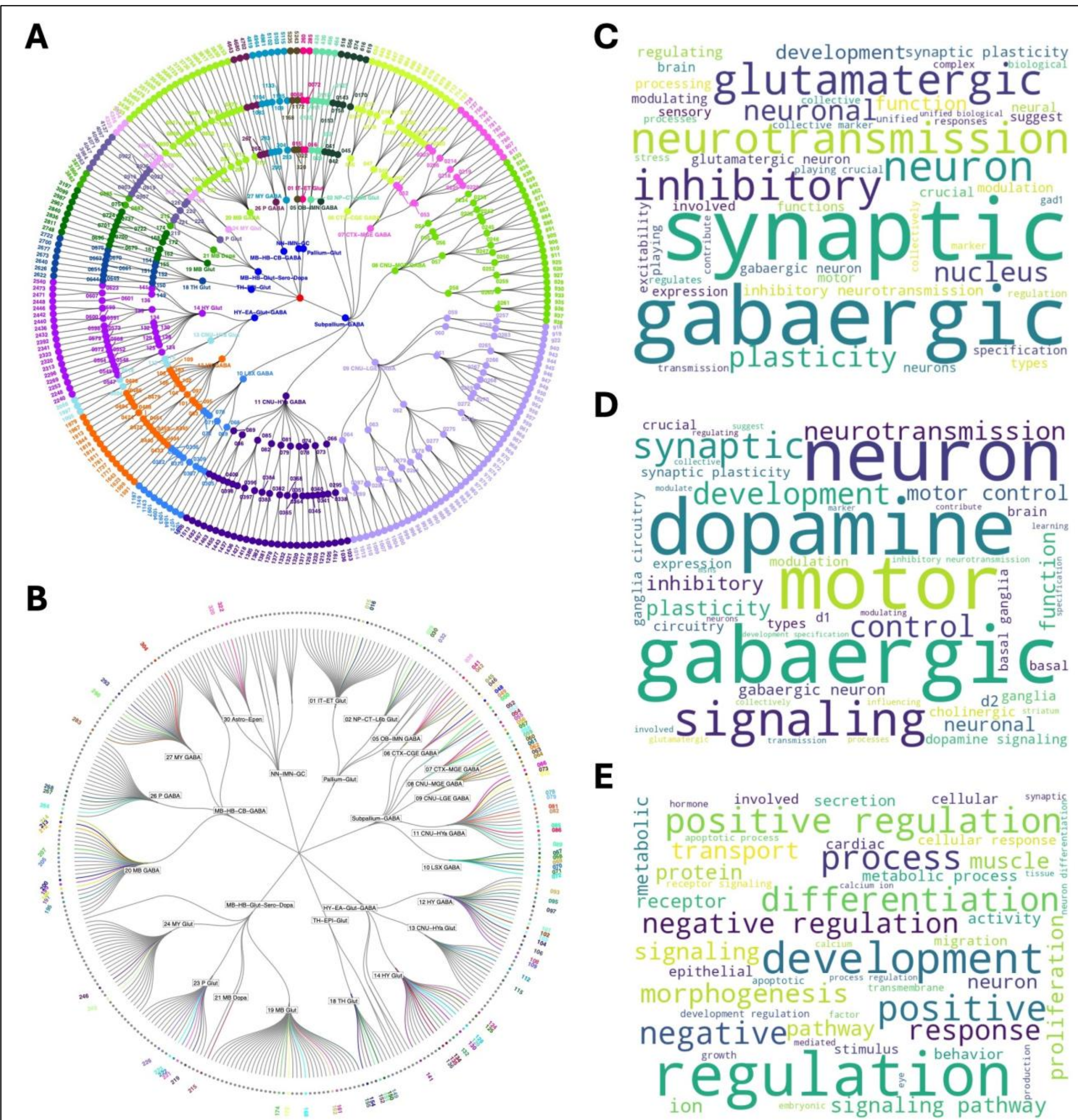

**Figure 4. BRAINCELL-AID enables rapid identification of brain cells with specific biological features or functions through a powerful search interface.** (A) Taxonomy structure of the mouse brain cell clusters associated with the keyword “Basal Ganglia”. (B) These “Basal Ganglia” clusters are mapped to the BICAN ABC Atlas taxonomy. (C) Word cloud of cell type annotations for all the mouse brain cell clusters generated by BRAINCELL-AID. (D) Word cloud of cell type annotation for mouse brain cell clusters with key word “Basal Ganglia” in their BRAINCELL-AID annotations. (E) Word cloud generated from GSEA annotation for the brain cell clusters with key word “Basal Ganglia” in their BRAINCELL-AID annotations, as we cannot find “Basal Ganglia” in any GSEA annotation.

noted—potentially indicating microglia–astrocyte interaction. Human Protein Atlas (HPA) data show that Slc1a3 is predominantly expressed in astrocytes, localized to endfeet that interface with blood vessels and other cells. This spatial context supports the LLM’s prediction of

microglia–astrocyte signaling and highlights Slc1a3's role in intercellular communication (Cluster 5312, merfish marker gene set, GPTON TopPM annotation).

Clusters 5313 and 5314 were annotated as BAMs, and BRAINCELL-AID effectively distinguished between homeostatic and activated states. Both clusters expressed core BAM markers (*Mrc1, Igf1, Csf1r*), supporting a homeostatic identity. However, Cluster 5314 also expressed activation-associated genes such as *Fos* and *Txnip*. Specifically, GPTON annotation captured that "Fos is involved in cell proliferation and differentiation in response to growth factors" with literature support (**Cluster 5314, TF marker gene set, GPTON TopPM annotation**). Cell Cluster Annotation Summary of this cluster specifically captured Txnip in regulating phagosomal acidification with literature support. These evidences suggest a transition of the cluster toward an inflammatory state as compared to Cluster 5313.[41] BRAINCELL-AID captured this subtle distinction in its annotation, highlighting the functional plasticity of BAMs within the CNS. This analysis confirms that BRAINCELL-AID produces meaningful, subtype-specific annotations that reflect both cell identity and activation state in glial populations.

Beyond refining annotations, BRAINCELL-AID introduces a major advancement: the ability to generate testable, neuroanatomically contextualized hypotheses. For example, spatial transcriptomic data for cluster 3854 revealed dopaminergic neurons expressing *Slc6a3*, *Gtf2a1l*, and *Aldh1a7* across the Substantia Nigra (SN), Ventral Tegmental Area (VTA), and Midbrain Raphe Nuclei (RAmb). Based on these patterns, BRAINCELL-AID inferred roles in dopaminergic signaling and potential GABA co-transmission—particularly in the RAmb, where such functionality remains poorly characterized (**Cluster 3854, cluster_combo marker gene set, GPTON TopGene annotation**).

This example illustrates BRAINCELL-AID's strength in both recovering known anatomy-function relationships and proposing novel functional hypotheses at the circuit level based on molecular signatures such as marker gene sets. By combining anatomical context with molecular signature and literature-informed reasoning, it enables the discovery of region- and cell-type-specific functional modules from large-scale transcriptomic datasets.

The system detects region-specific gene co-expression patterns and infers putative functions of gene ensembles. By anchoring these profiles within anatomical boundaries, it links molecular identity to neurotransmitter systems, cell types, and neural circuits. Importantly, BRAINCELL-AID goes beyond confirming known biology—it predicts novel combinatorial gene functions related to co-transmission, neuromodulatory coupling, and circuit-specific signaling.

These predictions serve as a foundation for scalable, data-driven hypothesis generation and experimental validation, grounded in anatomical precision. In sum, BRAINCELL-AID provides a robust framework for functional annotation and hypothesis generation, advancing the development of a comprehensive, expert-informed brain cell atlas across species.

**BRAINCELL-AID portal**

We developed the BRAINCELL-AID portal to host detailed annotations for brain cell types and their associated gene sets (**Figure 5; Methods**). This comprehensive and user-friendly resource enables in-depth exploration of brain cell taxonomy through annotated brain cell types and marker gene sets generated by our agent-based workflow. The portal features a fixed navigation bar for quick access to the Home and Contact pages.

Given that BICAN primarily uses single-cell RNA sequencing (scRNA-seq) to identify brain cell types, we prominently display all marker gene sets on the homepage for ease of access. Each marker gene set entry includes key taxonomic and gene information: Cluster ID, Super Type, Class Label, NT Type Label, Marker Type, and Marker Genes. A "View" button links directly to the full BRAINCELL-AID annotation (**Figure 5A**).

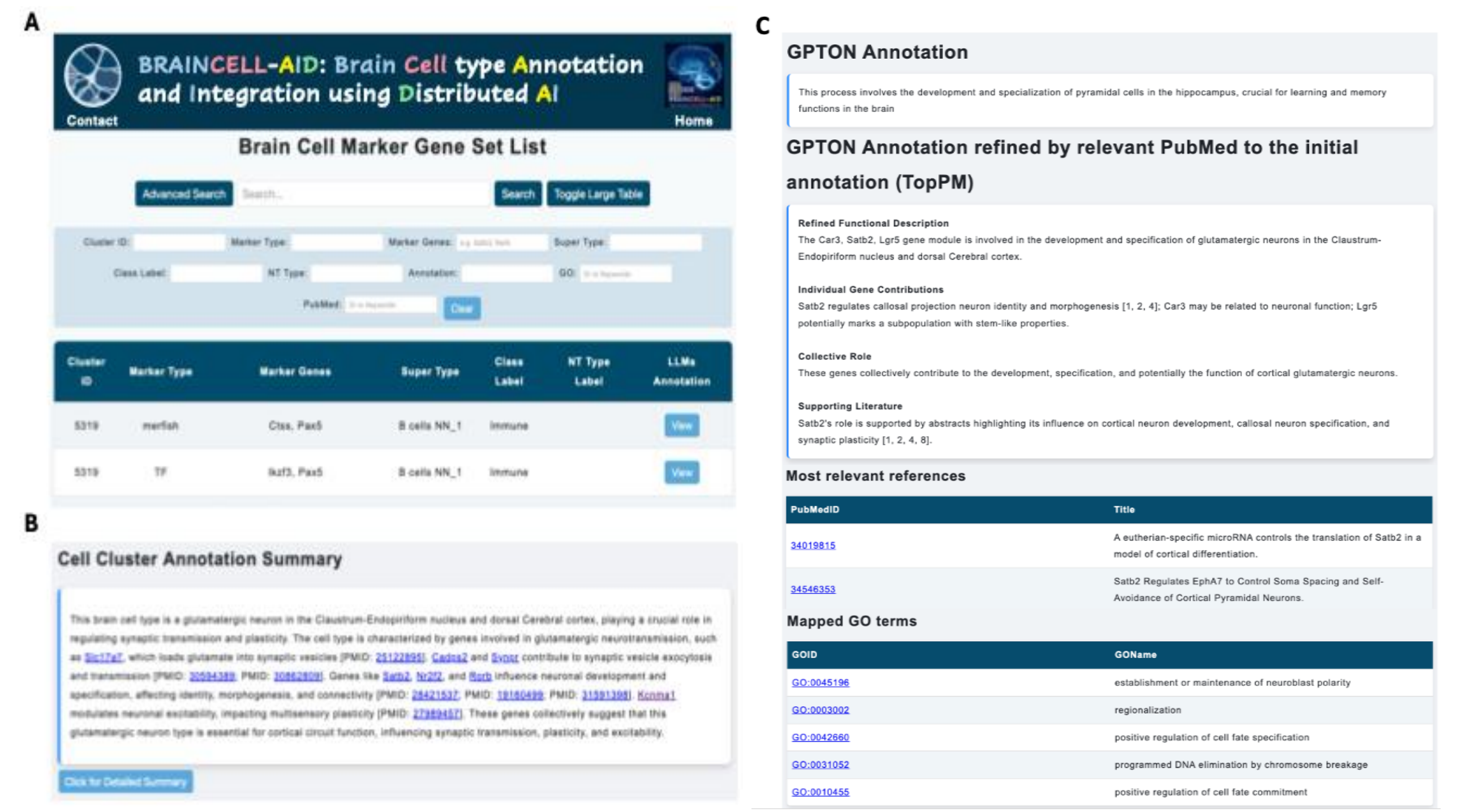

**Figure 5**. **Application as Web-based Community Resource.** (A) BRAINCELL-AID web portal homepage, where all marker gene sets are listed. Advanced search is provided for comprehensive search of BRAINCELL-AID annotation. (B) Screenshot of brain cell type annotation generated from summarizing all marker gene set annotations. Access to the detailed summary is also provided. (C) Shown partial Marker Gene Set annotation that includes the predicted annotation, mapped GO terms, TopPM annotation and a part of the relevant PubMed references for one cluster in the web-based repository. The webpage also includes the mapped GO terms for TopPM annotation, and all the corresponding information for TopGene annotation.

To enhance usability, the interface includes "Previous" and "Next" buttons for page-by-page browsing, page number shortcuts for faster navigation across the 21,275 annotated gene sets (spread over 1,064 pages), and a "Toggle Large Table" button to expand the view from 20 to 90 gene sets per page.

A simple search box enables quick search for marker gene sets using one of the six fields, while an Advanced Search feature supports robust queries of nine fields by Cluster ID, Gene Symbol, functional annotation terms, etc. The results link users directly to detailed gene cluster pages. Filtering options allow users to narrow results by hierarchical classification, marker type, or annotation source. These features further support data FAIRness by facilitating searches based on controlled taxonomy terms, GO annotations, and PubMed IDs (**Figure 5A**).

Each gene set has a dedicated page with rich metadata and annotations, accessible via the "View" button. The annotation is organized into three clearly defined sections:

1. **Brain Cell Cluster Information**
   This section presents taxonomy data from the ABC Atlas, including Cluster ID, Marker Type, Super Type, Class Label, and NT Type Combo Label. It also includes a link to the Brain Knowledge Platform for further context.
2. **Cell Cluster Annotation Summary (Figure 5B)**
   BRAINCELL-AID provides a concise summary of each cell type, including anatomical location, neurotransmitter identity (if available), and literature-based functional insights inferred from marker genes. Hyperlinked gene symbols and PubMed IDs allow quick access to supporting evidence. A "Click for Detailed Summary" button reveals a longer annotation, often including reasoning paragraphs explaining how LLM assembled the annotations.
3. **Marker Gene Set (Figure 5C)**
   This section includes details on marker type and gene members, along with multiple layers of annotation:
   - Original LLM-based annotation (GPTON Annotation)
   - RAG-enhanced annotation via TopPM and TopGene
   - GO term mappings for all annotations
   - Literature references linked to PubMed for RAG annotations

All GO terms and PubMed links are fully integrated to provide seamless access to supporting databases.

**Toward community-based brain cell type annotation**

By leveraging LLMs and AI agents, BRAINCELL-AID represents a foundational step toward the comprehensive annotation of brain cell types. Our approach—combining LLM fine-tuning with Retrieval-Augmented Generation (RAG)—significantly improves gene set annotation by grounding results in published literature. This leads to highly relevant annotations, capturing ground truth in 77% of mouse and 74% of human cases and laying the groundwork for engaging the neuroscience community in a collaborative effort to annotate, refine, and ultimately define brain cell types across species (**Figure 3E**).

To support community involvement, the BRAINCELL-AID portal enables users to review, evaluate, and submit corrections or additions to existing annotations by providing an "Annotate this cluster" button (**Figures 6A and 6B**). The Annotate button will enable functions that allow users to edit each gene cluster page's content as an interactive form for submitting updates (**Figures 6C and S6)**. User submissions are timestamped, stored alongside original entries, and displayed in parallel. This evolving annotation system promotes transparency, accuracy, and collective knowledge building.

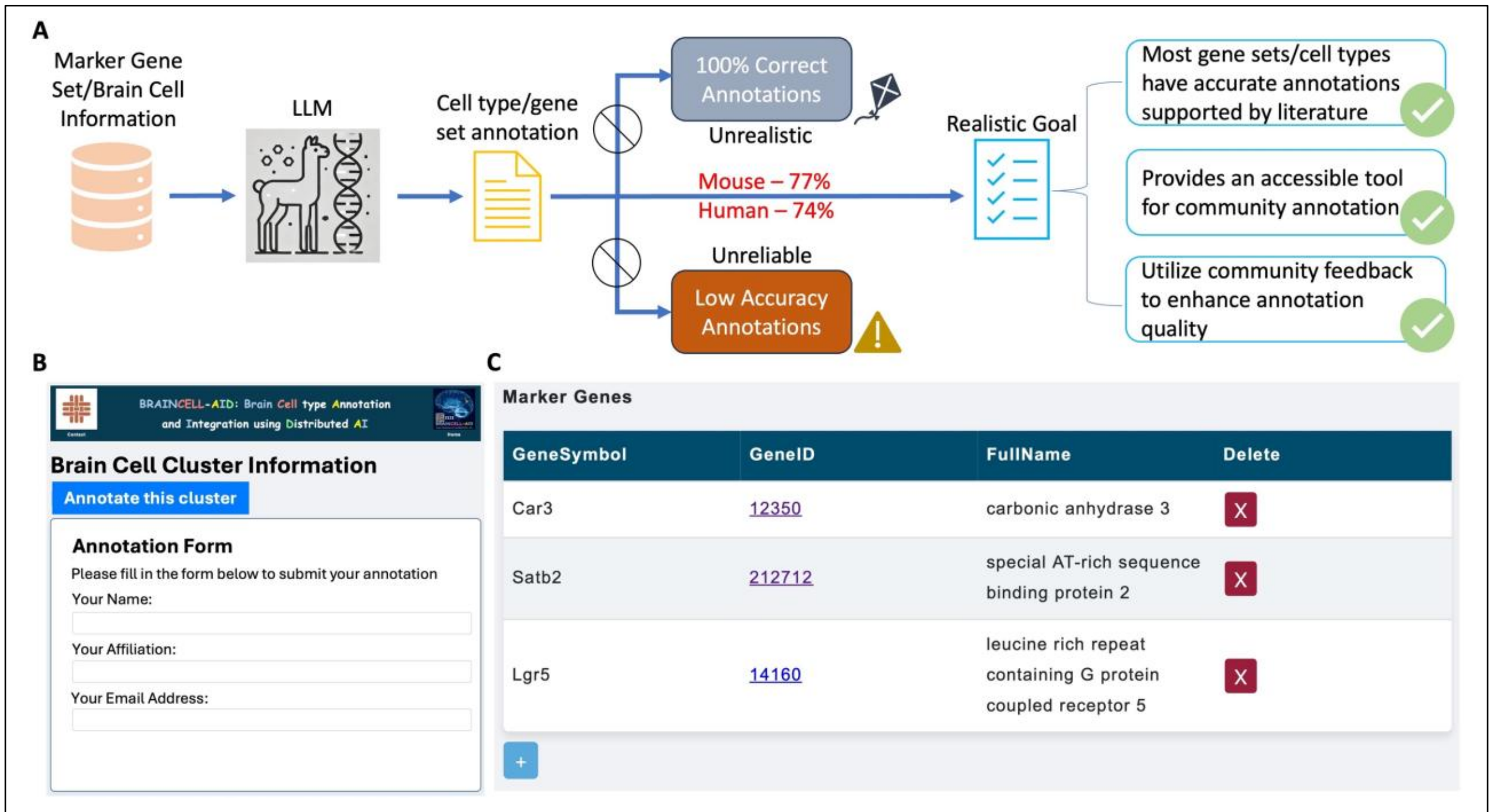

**Figure 6**. **BRAINCELL-AID for community**. (A) The illustration of a reasonable goal that our proposed network achieves to find a balance between practicality and usefulness for community annotation of brain cell types. (B) BRAINCELL-AID web portal allows users to input their information through an annotation form. (C) Editing feature of BRAINCELL-AID portal that allows users to add or delete marker genes. Other editing features are illustrated in Figure S6.

We envision this portal as a community-driven resource for exploring and refining annotations of emerging gene clusters and brain cell types. By integrating external references and offering seamless access to relevant databases, we empower domain experts to evaluate and validate annotations with ease.

We invite neuroscientists, molecular biologists, and computational experts to contribute their expertise directly through the portal. Together, this collaborative model fosters community ownership of the data and advances the development of a validated, species-spanning brain cell taxonomy grounded in both AI and expert insight.

## DISCUSSION

Understanding brain cell types is a fundamental step toward deciphering brain function, yet the complexity and heterogeneity of brain cells pose major challenges. Although scRNA-seq technologies have enabled the identification of transcriptional signatures at unprecedented resolution, annotating these gene sets remains challenging—especially for genes with limited or poorly defined functional annotations that are vague, generic, or superficial. To address this gap, BRAINCELL-AID leverages LLMs and AI agents to deliver a novel, scalable framework for gene set annotation and brain cell type definition.

A key innovation in our approach is GPTON, a strategy that adapts LLMs for biological reasoning by verbalizing GO terms. Rather than relying solely on structured ontologies, which are often too rigid or complex for LLMs to process directly, GPTON reframes ontology terms as natural language prompts. This allows the model to generate coherent, biologically relevant summaries and map them back to established GO terms. By fine-tuning LLMs with these verbalized terms, we enable accurate annotation even for gene sets lacking direct ontology labels or derived from novel experimental studies. This verbalization approach not only improves gene set annotation but also opens a pathway to infuse broader biological concepts into LLM reasoning for brain cell type annotation and definition.

Unlike traditional statistical methods such as GSEA, which rely on enrichment p-values, GPTON operates using machine learning-based inference. Built on the 70- and 405-billion-parameter Llama models, GPTON derives confidence from its training and internal representations rather than hypothesis testing. Its effectiveness is reflected in strong performance metrics—for example, nearly 70% of gene sets were annotated with at least one term highly relevant to ground truth within the top five predictions. This represents a significant advancement in the biological accuracy and generalizability of LLM-based annotation methods.

To further increase reliability and mitigate LLM hallucinations, we integrated RAG, which grounds annotations in biomedical literature. Our agentic workflow identifies relevant PubMed abstracts using semantic vector search and incorporates them into the annotation process. This is particularly critical for novel gene sets, where existing labels may be absent. The RAG-enhanced system significantly improved annotation accuracy—achieving 77% and 74% relevance to ground truth in mouse and human datasets, respectively—demonstrating the utility of literature-grounded AI agents for functional genomics. By automating evidence retrieval and supporting transparency in annotation, RAG adds both robustness and interpretability to the framework.

The flexibility of our approach enables rich annotation of gene sets with limited or no existing ontology, including those emerging from new experimental contexts. This versatility suggests broader applications of the framework beyond gene set annotation. In particular, our system could be extended to human disease research and drug discovery—helping identify disease-associated genes, map functional pathways, and prioritize therapeutic targets through literature-supported predictions.[42,43]

BRAINCELL-AID annotation also helps to addresses the question, “Why are there so many brain cell clusters/types?” The 5,332 clusters identified in the adult mouse brain indeed represent an overwhelming number. Detailed examination of Clusters 5313 and 5314 illustrates the underlying complexity: both are BAMs, yet Cluster 5314 expresses activation-associated genes such as *Fos* and *Txnip*, indicating a transition toward an inflammatory state. This finding suggests that some clusters may represent the same cell type in different states—such as inflammatory, developmental, or others—rather than entirely distinct cell types. BRAINCELL-AID annotation effectively detects these subtle distinctions, yielding novel insights into the neurobiology of brain cells. It demonstrates the high discriminative power of molecular profiling, the foundation of the BICCN and BICAN approach, which enables unprecedented resolution in identifying cell types and states. When integrated with large language models and AI agent–based annotation, this approach provides deep biological insights into brain cell function beyond the reach of previous methods, advancing the overarching goals of the BRAIN Initiative.

To support community-driven knowledge building, we developed the BRAINCELL-AID web portal, which integrates fine-tuned LLM outputs, literature references, and RAG-refined annotations. The portal allows users to review, suggest corrections, and contribute new insights for each brain cell cluster, fostering a collaborative model for continuous annotation improvement. With community engagement, this platform can evolve into a curated, literature-supported knowledgebase of brain cell types—complementing resources such as the ABC Atlas[3,5] and serving as a foundational tool for neuroscience.

Looking ahead, the performance of BRAINCELL-AID can be further enhanced by expanding its training corpus, incorporating more structured biological knowledge, and refining the Query Agent. These improvements would not only advance gene set annotation but also extend the system's impact across biomedical domains. For example, future applications could include mechanistic studies of human diseases, exploration of gene regulatory networks, and identification of context-specific drug targets. Moreover, BRAINCELL-AID generates testable hypotheses that can be evaluated in laboratory experiments, offering a path to experimental validation and deeper functional understanding.

In conclusion, BRAINCELL-AID introduces a powerful, LLM- and AI agent-based framework for functional annotation of brain cells. By combining ontology verbalization, literature grounding, and agentic workflows, it sets a new standard for the integration of structured biological knowledge into language models. Its modularity, accuracy, and scalability make it a promising tool for accelerating discovery in neuroscience and beyond.

## RESOURCE AVAILABILITY

### Lead contact

For correspondence and data request, please contact W. Jim Zheng (wenjin.j.zheng@uth.tmc.edu).

### Material availability

N/A

### Data availability

All the BRAINCELL-AID generated data are provided through its web portal at: https://biodataai.uth.edu/BRAINCELL-AID.

The data used in this manuscript were all downloaded from publicly available data sources. Specifically, the GO term annotation information regarding the human gene sets was downloaded from the MSigDB website (https://www.gsea-msigdb.org/gsea/msigdb/collections.jsp). Similarly, the MSigDB website also contains the GO term annotation regarding the mouse gene sets (https://www.gsea-msigdb.org/gsea/msigdb/mouse/collections.jsp). The detailed information regarding each gene (gene_info) was collected from the NCBI (RRID:SCR_006472) Gene FTP database (https://ftp.ncbi.nih.gov/gene/DATA/). The gene2pubmed file was downloaded from the NCBI FTP database (https://ftp.ncbi.nlm.nih.gov/gene/DATA/gene2pubmed.gz). The PubMed (RRID:SCR_004846) abstracts were downloaded from the NCBI FTP database

(https://ftp.ncbi.nlm.nih.gov/pubmed/). The brain cell expression data were downloaded from ABC Atlas.

**ACKNOWLEDGEMENTS**
Rongbin Li is supported by NLM Training Program in Biomedical Informatics & Data Science for Predoctoral & Postdoctoral Fellows under the Gulf Coast Consortia grant 5T15LM007093-31. Jinbo Li and Meaghan Ramlakhan were supported by the Cancer Prevention and Research Institute of Texas training grant RP210045. This work is also partly supported by the National Institutes of Health (NIH) through grants 1UL1TR003167, 1UM1TR004906-01, 1R01AG066749, 1U24MH130988-01, 5R56AG069880-02, Department of Defense W81XWH-22-1-0164, 1R01AR081280-01A1, 5R01AG074283-04, and the Cancer Prevention and Research Institute of Texas through grant RP170668 (WJZ).

**AUTHOR CONTRIBUTIONS**
WJZ, MH, and HXu developed the strategic vision of the BICAN AI Initiative within the BICAN Consortium, where this work is a pilot project; WJZ, ZL conceptualized the GPTON and AI Agent strategies of the project; RL, WC, ZL, RMC, ZWu, WJZ developed the methodology; RMC, WC, NM, ZWu, ZWise investigated scientific quality of BRAINCELL-AID annotation; RL, WC, JL, HXing, ZL, WJZ conducted investigation for ontology verbalization; RL, JL implemented AI agent system; WC conducted Word Cloud analysis and GSEA; RL, ZL, AS, NM, WC, WJZ visualized the results; NJ advised on BG data analysis; JL, MR, HH developed web portal; WJZ, ZL supervised the entire project; RL, WC, ZL, WJZ drafted the original manuscript; RL, WC, HXu, MH, WJZ reviewed and finalized the manuscript.

**REFERENCES**
1 Zeggini, E. *et al.* Meta-analysis of genome-wide association data and large-scale replication identifies additional susceptibility loci for type 2 diabetes. *Nat Genet* **40**, 638-645 (2008). https://doi.org:10.1038/ng.120
2 Consortium, E. P. *et al.* Identification and analysis of functional elements in 1% of the human genome by the ENCODE pilot project. *Nature* **447**, 799-816 (2007). https://doi.org:10.1038/nature05874
3 Yao, Z. *et al.* A high-resolution transcriptomic and spatial atlas of cell types in the whole mouse brain. *Nature* **624**, 317-332 (2023). https://doi.org:10.1038/s41586-023-06812-z
4 Collins, F. S. *et al.* New goals for the U.S. Human Genome Project: 1998-2003. *Science* **282**, 682-689 (1998). https://doi.org:10.1126/science.282.5389.682
5 Siletti, K. *et al.* Transcriptomic diversity of cell types across the adult human brain. *Science* **382**, eadd7046 (2023). https://doi.org:10.1126/science.add7046
6 Subramanian, A. *et al.* Gene set enrichment analysis: a knowledge-based approach for interpreting genome-wide expression profiles. *Proc Natl Acad Sci U S A* **102**, 15545-15550 (2005). https://doi.org:10.1073/pnas.0506580102
7 Cui, H. *et al.* scGPT: toward building a foundation model for single-cell multi-omics using generative AI. *Nature methods* **21**, 1470-1480 (2024).
8 Xiao, Y. *et al.* Cellagent: An llm-driven multi-agent framework for automated single-cell data analysis. *arXiv preprint arXiv:2407.09811* (2024).
9 Godec, P., Zupan, B., Tanko, V. & Stražar, M. in *2022 IEEE 10th International Conference on Healthcare Informatics (ICHI).* 633-638 (IEEE).
10 Consortium*, T. T. S. *et al.* The Tabula Sapiens: A multiple-organ, single-cell transcriptomic atlas of humans. *Science* **376**, eabl4896 (2022).
11 Perez, R. K. *et al.* Single-cell RNA-seq reveals cell type–specific molecular and genetic associations to lupus. *Science* **376**, eabf1970 (2022).

12. Ashburner, M. *et al.* Gene ontology: tool for the unification of biology. The Gene Ontology Consortium. *Nat Genet* **25**, 25-29 (2000). https://doi.org:10.1038/75556
13. Gene Ontology, C. *et al.* The Gene Ontology knowledgebase in 2023. *Genetics* **224** (2023). https://doi.org:10.1093/genetics/iyad031
14. Consortium, G. O. Expansion of the Gene Ontology knowledgebase and resources. *Nucleic acids research* **45**, D331-D338 (2017).
15. Consortium, G. O. The gene ontology resource: 20 years and still GOing strong. *Nucleic acids research* **47**, D330-D338 (2019).
16. Joachimiak, M. P., Caufield, J. H., Harris, N. L., Kim, H. & Mungall, C. J. Gene set summarization using large language models. *ArXiv*, arXiv: 2305.13338 v13333 (2024).
17. AI, M. *Introducing LLaMA 4: Advancing Multimodal Intelligence*, <https://ai.meta.com/blog/llama-4-multimodal-intelligence/> (2024).
18. Li, M., Kilicoglu, H., Xu, H. & Zhang, R. Biomedrag: A retrieval augmented large language model for biomedicine. *Journal of Biomedical Informatics* **162**, 104769 (2025).
19. Thirunavukarasu, A. J. *et al.* Large language models in medicine. *Nature medicine* **29**, 1930-1940 (2023).
20. Taylor, R. *et al.* Galactica: A large language model for science. *arXiv preprint arXiv:2211.09085* (2022).
21. Gu, Y. *et al.* Domain-specific language model pretraining for biomedical natural language processing. *ACM Transactions on Computing for Healthcare (HEALTH)* **3**, 1-23 (2021).
22. Alrowili, S. & Vijay-Shanker, K. in *Proceedings of the 20th workshop on biomedical language processing.* 221-227.
23. Li, Z. *et al.* Ensemble pretrained language models to extract biomedical knowledge from literature. *Journal of the American Medical Informatics Association* **31**, 1904-1911 (2024).
24. Babaei Giglou, H., D'Souza, J. & Auer, S. in *International Semantic Web Conference.* 408-427 (Springer).
25. Caufield, J. H. *et al.* Structured prompt interrogation and recursive extraction of semantics (SPIRES): A method for populating knowledge bases using zero-shot learning. *Bioinformatics* **40**, btae104 (2024).
26. He, Y., Chen, J., Dong, H. & Horrocks, I. Exploring large language models for ontology alignment. *arXiv preprint arXiv:2309.07172* (2023).
27. Lewis, P. *et al.* Retrieval-augmented generation for knowledge-intensive nlp tasks. *Advances in neural information processing systems* **33**, 9459-9474 (2020).
28. Ament, S. A. *et al.* The Neuroscience Multi-Omic Archive: A BRAIN Initiative resource for single-cell transcriptomic and epigenomic data from the mammalian brain. *bioRxiv*, 2022.2009.2008.505285 (2022). https://doi.org:10.1101/2022.09.08.505285
29. Kenney, M. *et al.* The Brain Image Library: A Community-Contributed Microscopy Resource for Neuroscientists. *Sci Data* **11**, 1212 (2024). https://doi.org:10.1038/s41597-024-03761-8
30. *Distributed Archives for Neurophysiology Data Integration (RRID: SCR_017571)*, <https://dandiarchive.org> (
31. Wooldridge, M. J. *An introduction to multiagent systems*. 2nd edn, (John Wiley & Sons, 2009).
32. Russell, S. J. & Norvig, P. *Artificial intelligence : a modern approach*. Fourth edition. edn, (Pearson, 2021).
33. Wang, Z. *et al.* GeneAgent: self-verification language agent for gene-set analysis using domain databases. *Nat Methods* **22**, 1677-1685 (2025). https://doi.org:10.1038/s41592-025-02748-6
34. Hu, M. *et al.* Evaluation of large language models for discovery of gene set function. *Nat Methods* **22**, 82-91 (2025). https://doi.org:10.1038/s41592-024-02525-x
35. Liberzon, A. *et al.* Molecular signatures database (MSigDB) 3.0. *Bioinformatics* **27**, 1739-1740 (2011). https://doi.org:10.1093/bioinformatics/btr260
36. Achiam, J. *et al.* Gpt-4 technical report. *arXiv preprint arXiv:2303.08774* (2023).
37. Kaljurand, K. & Fuchs, N. E. Verbalizing owl in attempto controlled english. (2007).

38 Li, R. *et al.* GPTON: Generative Pre-trained Transformers enhanced with Ontology Narration for accurate annotation of biological data. *arXiv preprint arXiv:2410.10899* (2024).

39 Lin, C.-Y. in *Text summarization branches out.* 74-81.

40 Gerrits, E., Heng, Y., Boddeke, E. W. G. M. & Eggen, B. J. L. Transcriptional profiling of microglia; current state of the art and future perspectives. *Glia* **68**, 740-755 (2020). https://doi.org:10.1002/glia.23767

41 Tsubaki, H., Tooyama, I. & Walker, D. G. Thioredoxin-Interacting Protein (TXNIP) with Focus on Brain and Neurodegenerative Diseases. *International Journal of Molecular Sciences* **21**, 9357 (2020). https://doi.org:10.3390/ijms21249357

42 Dahl, S. G., Kristiansen, K. & Sylte, I. Bioinformatics: from genome to drug targets. *Annals of medicine* **34**, 306-312 (2002).

43 Maleki, F., Ovens, K., Hogan, D. J. & Kusalik, A. J. Gene set analysis: challenges, opportunities, and future research. *Frontiers in genetics* **11**, 654 (2020).